\documentclass[runningheads]{llncs}
\usepackage{graphicx}

\usepackage{tikz}
\usepackage{comment}
\usepackage{amsmath,amssymb} 
\usepackage{color}
\usepackage{bbm}
\usepackage{caption}
\usepackage{float}
\usepackage{booktabs}
\usepackage{comment}
\usepackage{orcidlink}

\usepackage{caption}
\usepackage{subcaption}

\usepackage[accsupp]{axessibility}  
\usepackage{xcolor}

\newcommand{\todo}[1]{\textcolor{red}{(#1)}}
\newcommand{\strawman}[1]{}

\begin{document}
\raggedbottom
\pagestyle{headings}
\mainmatter
\def\ECCVSubNumber{186}  





\title{Improving Contrastive Learning on Visually Homogeneous Mars Rover Images}




\titlerunning{Improving Contrastive Learning for Mars}
%
\author{Isaac Ronald Ward\inst{1,3}\orcidlink{0000-0002-3418-3138} \and
Charles Moore\inst{2} \and 
Kai Pak\inst{1}\orcidlink{0000-0002-2816-0167} \and\\
Jingdao Chen\inst{2} \and
Edwin Goh\inst{1}}
\authorrunning{I. R. Ward et al.}
%
\institute{
    Jet Propulsion Laboratory, California Institute of Technology\\
    \and
    Computer Science and Engineering, Mississippi State University\\
    \and
    Department of Computer Science, University of Southern California\\
    \email{isaac.r.ward@jpl.nasa.gov},
    \email{cam1271@msstate.edu},
    \email{kai.pak@jpl.nasa.gov},
    \email{chenjingdao@cse.msstate.edu},
    \email{edwin.y.goh@jpl.nasa.gov}, 
}

\maketitle

\begin{abstract} 
Contrastive learning has recently demonstrated superior performance to supervised learning, despite requiring no training labels. We explore how contrastive learning can be applied to hundreds of thousands of unlabeled Mars terrain images, collected from the Mars rovers Curiosity and Perseverance, and from the Mars Reconnaissance Orbiter. Such methods are appealing since the vast majority of Mars images are unlabeled as manual annotation is labor intensive and requires extensive domain knowledge. Contrastive learning, however, assumes that any given pair of distinct images contain distinct semantic content. This is an issue for Mars image datasets, as any two pairs of Mars images are far more likely to be semantically similar due to the lack of visual diversity on the planet's surface. Making the assumption that pairs of images will be in visual contrast --- when they are in fact not --- results in pairs that are falsely considered as negatives, impacting training performance. In this study, we propose two approaches to resolve this: 1) an unsupervised deep clustering step on the Mars datasets, which identifies clusters of images containing similar semantic content and corrects false negative errors during training, and 2) a simple approach which mixes data from different domains to increase visual diversity of the total training dataset. Both cases reduce the rate of false negative pairs, thus minimizing the rate in which the model is incorrectly penalized during contrastive training. These modified approaches remain fully unsupervised end-to-end. To evaluate their performance, we add a single linear layer trained to generate class predictions based on these contrastively-learned features and demonstrate increased performance compared to supervised models; observing an improvement in classification accuracy of $3.06\%$ using only $10\%$ of the labeled data.


\end{abstract}

\section{Introduction} 
\strawman{
\begin{itemize}
    \item Establish territory (scope)
    \item Establish niche (scope)
    \item Occupy niche (scope)
    \item What field does this research belong to?
    \item What particular problem area have I focused on?
    \item What scienfitic question do I aim to answer?
    \item Why is it important?
    \item In what sense will I advance our knowledge about it?
    \item Where are we?
    \item Where are we going?
    \item Why do we want to go there?
    \item Really need to comment on the false negative problem early.
    \item What is THE one research question?
\end{itemize}
}

The primary goal for exploring Mars is to collect data pertaining to the planet's geology and climate, identify potential biological markers to find evidence of past or present life, and study the planet in preparation for eventual human exploration \cite{nasa2022marsgoals}. As it is currently infeasible for humans to do this work, autonomous rovers have emerged as the primary means to explore the Martian surface and collect images and other data. The deluge of images produced from these rovers --- on the order of hundreds of thousands of images --- has provided the opportunity to apply deep learning (DL) based computer vision methods to tackle a variety of science and engineering challenges.

Existing DL-based approaches have tackled the aforementioned challenge using supervised (transfer) learning and typically require thousands of annotated images to achieve reasonable performance \cite{wagstaff2018deep,wagstaff2021mars}. Although a manual labeling approach may seem viable, Mars images contain subtle class differences and fine-grained geological features that require highly specialized scientific knowledge and expertise, meaning that scaling manual efforts to the volume of required data is difficult, if not impractical. Several efforts at training citizen scientists to annotate such datasets have been met with some success \cite{sprinks2016mars}. However, these logistically complex efforts can introduce inconsistencies that must be resolved by experts \cite{swan2021ai4mars} to prevent ambiguities/bias.

Self-supervised learning (SSL) can help to circumvent the need for large-scale labeling efforts, since these techniques do not require labeled data to train.
Contrastive learning (CL) --- a type of SSL --- has demonstrated much success in this domain in the past few years, and continues to gain research momentum thanks to the promise of leveraging unlabeled images to achieve state-of-the-art performance on large-scale vision benchmarks \cite{chen2020simclr}. These techniques generally create psuedo-labels by leveraging the intrinsic properties of images, such as prior knowledge that two views (augmented crops) originating from the same source image must belong to the same semantic class. These pseudo-labels are then used in conjunction with a contrastive loss function to help the model learn representations that attract similar (positive) samples and repel different (negative) samples, and in doing so, identify an image's defining visual features. 

The contrastive loss function implicitly assumes that an image and it's views define a unique semantic classes. 
While this implicit assumption (herein referred to as the \textit{contrastive assumption}) may be practical in diverse vision datasets with a large number of balanced classes, it is problematic for planetary science applications due to the homogeneity of planetary images. This, in combination with small batch sizes, increases the probability that multiple source images within the same batch are semantically similar, which further invalidates this assumption. Mars rover images, for example, are regularly taken immediately after one another, and $90\%$ of their content may be overlapping. 
This results in the incorrect assignment of pseudo-labels, diminishing performance overall. \textbf{In short, any two images from a Mars rover dataset may not necessarily provide the contrast required for effective contrastive learning.}

In this work, we demonstrate how contrastive learning can be used to extract useful features from Mars rover images without the need for labels. We also explore how simple modifications to the contrastive learning process can improve performance when the contrastive assumption is violated, without introducing the need for any human supervision. Our key contributions are thus:

\begin{enumerate}
    \item An improved method of contrastive training with a cluster-aware approach that improves the contrastive loss formulation.
    \item Evidence that mixed-domain datasets can markedly improve the performance of downstream vision tasks by increasing semantic diversity.
    \item Demonstrations of how SSL followed by supervised fine-tuning/linear evaluation with limited labels can exceed the performance of published supervised baselines on Mars-related vision tasks.
\end{enumerate}

\section{Related work}
\label{sec:related_work}

\textbf{Supervised learning for Mars images}. Wagstaff et al. proposed a fully supervised approach to training AlexNet-based \cite{krizhevsky2012alexnet} Mars classification models \cite{wagstaff2018deep,wagstaff2021mars}. In these works, benchmark datasets were created to validate the classification performance of the trained models: the Mars Science Laboratory (MSL) dataset, and the High Resolution Imaging Science Experiment (HiRISE) dataset. In this work, we use these datasets to benchmark and compare the downstream task performance of our trained feature extractors (more details in Section \ref{sec:datasets}). Other supervised approaches have since improved on the initial results reported by Wagstaff et al. by using attention-based models \cite{chakravarthy2021mrscatt}.





\textbf{Semi-supervised learning for Mars images}. Wang et al. engineered a semi-supervised learning approach tailored for the semantic content of Mars rover images \cite{wang2021semi}. Their approach ignores problematic (redundant) training samples encountered during contrastive learning by making use of labels. Their proposed multiterm loss function contains both supervised and unsupervised terms, thus creating a semi-supervised approach.



\textbf{Self-supervised learning for Mars Images}. Panambur et al. extract granular geological terrain information without the use of labels for the purpose of clustering sedimentary textures encountered in $30,000$ of Curiosity's Mast camera (Mastcam) images \cite{panambur2022self}. They modify a neural network architecture that was originally designed for texture classification so that it can support self-supervised training, and use a metric learning objective that leverages triplet loss to generate a learning signal. The K-nearest neighbors (KNN) algorithm is then used on the embeddings to cluster, and thus support the querying of the data. The results of this deep clustering are validated by planetary scientists, and a new taxonomy for geological classification is presented.



\textbf{Self-supervised learning for Earth observations}. Learning based approaches for Earth observation predictions have long been of interest to the space community \cite{kucik2021snn,helber2019eurosat}. Wang et al. introduced the `Self-Supervised Learning for Earth Observation, Sentinel-1/2' (SSL4EO-S12) dataset and illustrate how SSL can be used to achieve comparable or superior performance to fully supervised counterparts \cite{wang2022ssl4eo}. This work uses techniques such as MoCo \cite{he2019moco,chen2020mocov2} and DINO \cite{caron2021emerging}, the former of which considers contrastive learning as a dictionary lookup problem, and builds a dynamic dictionary solution that leverages a queue with a moving-averaged encoder, and the latter of which leverages the properties of vision transformers \cite{dosovitskiy2020image} trained with large batch sizes. 



\textbf{Contrastive learning for Mars images}. A Simple Framework for Contrastive Learning (SimCLR, proposed in \cite{chen2020simclr} and improved upon in \cite{chen2020simclrv2}) has been used to train discriminant and performant feature extractors in a self-supervised manner across many domains. In \cite{goh2022mars}, a deep segmentation network is pretrained on unlabeled images using SimCLR and trained further in a supervised manner on a limited set of labeled segmentation data (only $161$ images). This approach outperforms fully supervised learning approaches by $2 - 10\%$.




\textbf{Relevant improvements to contrastive learning}. A number of approaches to detect and counteract issues relating to violating the contrastive assumption are outlined in \cite{huynh2022boosting,chen2021incremental}. Such techniques generate a support set of views for any given image, and use cosine similarities between the support set images and an incoming view to detect if the potential pair will be a false negative. Techniques that use this manner of false negative detection have demonstrably outperformed standard contrastive learning techniques on the ImageNet dataset \cite{russakovsky2015imagenet}. 



\section{Datasets}
\label{sec:datasets}

\strawman{
\begin{itemize}
    \item Because we're dealing with unsupervised / self-supervised learning, we have
    \item Unsupervised datasets
    \item Supervised datasets (same domain)
    \item Out of domain supervised datasets because we want to test performance there too (generalisation of method)
    \item Downstream tasks are classification but could conceivably use extracted features for any downstream task
\end{itemize}
}


\begin{figure}[htbp]
    \centering
    \includegraphics[width=0.8\textwidth]{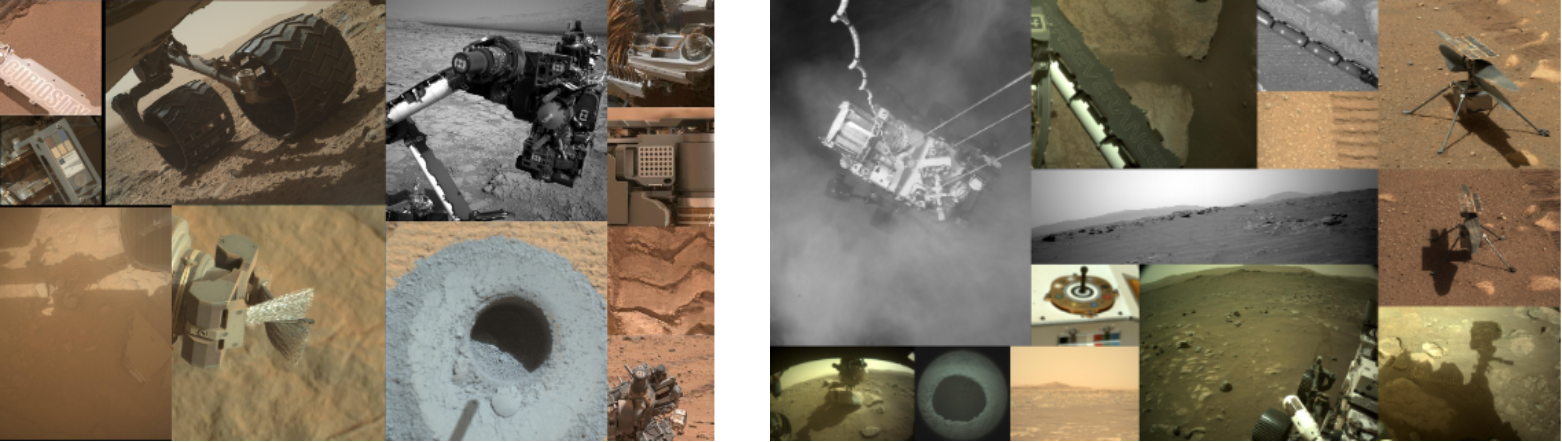}
    \caption*{(a) Unlabeled Curiosity images (left), and (b) unlabeled Perseverance images (right)}
    
    \includegraphics[width=0.8\textwidth]{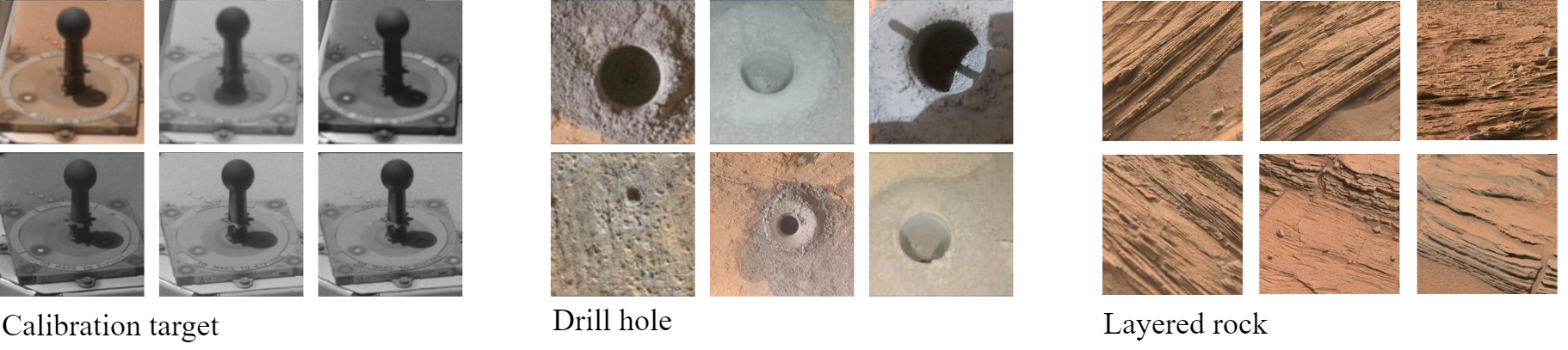}
    \caption*{(c) Labeled MSL v2.1 images \cite{wagstaff2018deep,wagstaff2021mars}, with $3$ of the $19$ classes shown.}
    
    \includegraphics[width=0.8\textwidth]{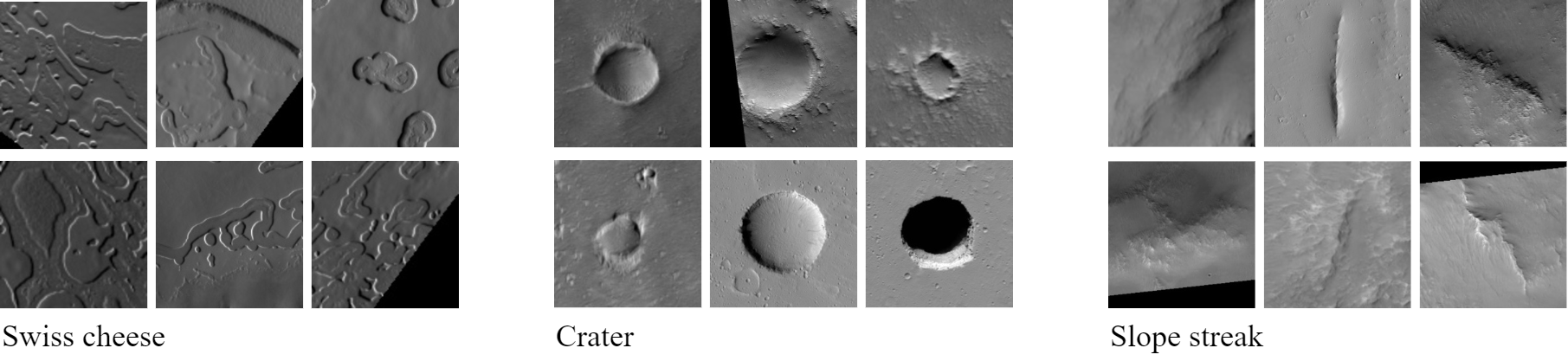}
    \caption*{(d) Labeled HiRISE v3.2 images \cite{wagstaff2018deep,wagstaff2021mars}, with $3$ of the $8$ classes shown.}
    
    \caption{The various datasets used in this work. The two unlabeled datasets (a and b) are entirely unstructured, but can be leveraged by contrastive learning techniques for training feature extractors. The two labeled datasets (c and d) are used for linear evaluation and benchmarking. We note that three of the four datasets are in the domain of Mars rover images (a, b, and c), whereas the HiRISE v3.2 dataset (d) is in the domain of orbital images. Further outlined in Table~\ref{table:datasets}.}
    \label{fig:datasets}
\end{figure}


\begin{table}[h]
  \setlength{\tabcolsep}{8pt}
  \caption{Statistics and descriptions of the datasets used in this work.}
  \label{table:datasets}
  \centering
  \begin{tabular}{p{1.6cm} p{6cm} p{1.7cm} l}
    \toprule
    Dataset & Description & Images (or & Labels \\
    name &  & train/test/val.) &  \\
    \midrule
    Perseverance    & Mars terrain images captured from the Perseverance rover across $19$ cameras. & 112,535 & No  \\
    Curiosity       & Mars terrain images captured from the Curiosity rover across $17$ cameras. & 100,000 & No  \\
    MSL v2.1 \cite{wagstaff2018deep,wagstaff2021mars}        & Mars terrain images captured from the Curiosity rover, partitioned into $19$ classes. & 5920 / 300 / 600 & Yes  \\
    HiRISE v3.2 \cite{wagstaff2018deep,wagstaff2021mars}     & Orbital images of Mars captured from the Mars Reconnaissance Orbiter, partitioned into $8$ classes. & 6997 / 2025 / 1793 & Yes  \\
    \bottomrule
  \end{tabular}
\end{table} 


\subsection{Perseverance rover images}

The primary source of training data in this work are unlabeled images from the Perseverance rover. We downloaded $112,535$ Perseverance rover images, captured during its traversal of the planet Mars from Sols 10 --- 400. These images can be accessed through NASA's Planetary Data System (PDS)\footnote{\url{https://pds.nasa.gov/}}. Each image is accompanied by metadata that outlines which of the rover's $19$ different cameras was used to take the image, the time of capture, and more. This results in a set of images that vary in resolution, colour profile, and semantic content. Image contents include the Mars surface, landscapes, terrain features, geologic features, the night sky, astronomical targets, calibration targets, the Perseverance rover itself, and more (as seen in Figure \ref{fig:datasets}).

\subsection{Curiosity rover images}



We consider two sets of Curiosity rover images in this work; an unlabeled set, and a labeled set. The unlabeled set consists of $100,000$ raw rover images, and is similar in semantic content to the Perseverance rover images. This dataset is used for contrastive learning in the same domain as the Perseverance rover images. The labeled dataset (herein referred to as MSL v2.1) was compiled and annotated by Wagstaff et al.'s \cite{wagstaff2018deep,wagstaff2021mars} and consists of $6820$ images divided into $19$ classes of interest.

The MSL v2.1 dataset is further divided by Sol into train, validation, and test subsets (see Table~\ref{table:datasets}). We use MSL v2.1's testing subset to benchmark the performance of the feature extraction networks that we train with contrastive learning, thus permitting the comparison of our results with the fully supervised approaches outlined in Wagstaff et al.'s work. We note that the MSL v2.1 dataset is imbalanced: one of the $19$ classes (entitled `nearby surface') accounts for $34.76\%$ of the images, whereas the `arm cover' class accounts for only $0.34\%$ of the images.




\subsection{Mars Reconnaissance Orbiter images}

We additionally use a labeled dataset of Mars images taken from the Mars Reconnaissance Orbiter's HiRISE camera, again compiled and annotated by Wagstaff et al. and split into train, validation, and test subsets (see Table~\ref{table:datasets}). This dataset (herein referred to as HiRISE v3.2) is again used entirely for benchmarking, but instead focuses on a separate visual domain (orbital images of Mars rather than rover-based images of Mars). Similarly to MSL v2.1, the HiRISE v3.2 dataset is imbalanced: one of the $8$ classes (entitled `other') accounts for $81.39\%$ of the images, whereas the `impact ejecta' class accounts for only $0.68\%$ of the images.



\section{Methods}
\label{sec:methods}

\begin{figure}
    \centering
    \includegraphics[width=0.9\textwidth]{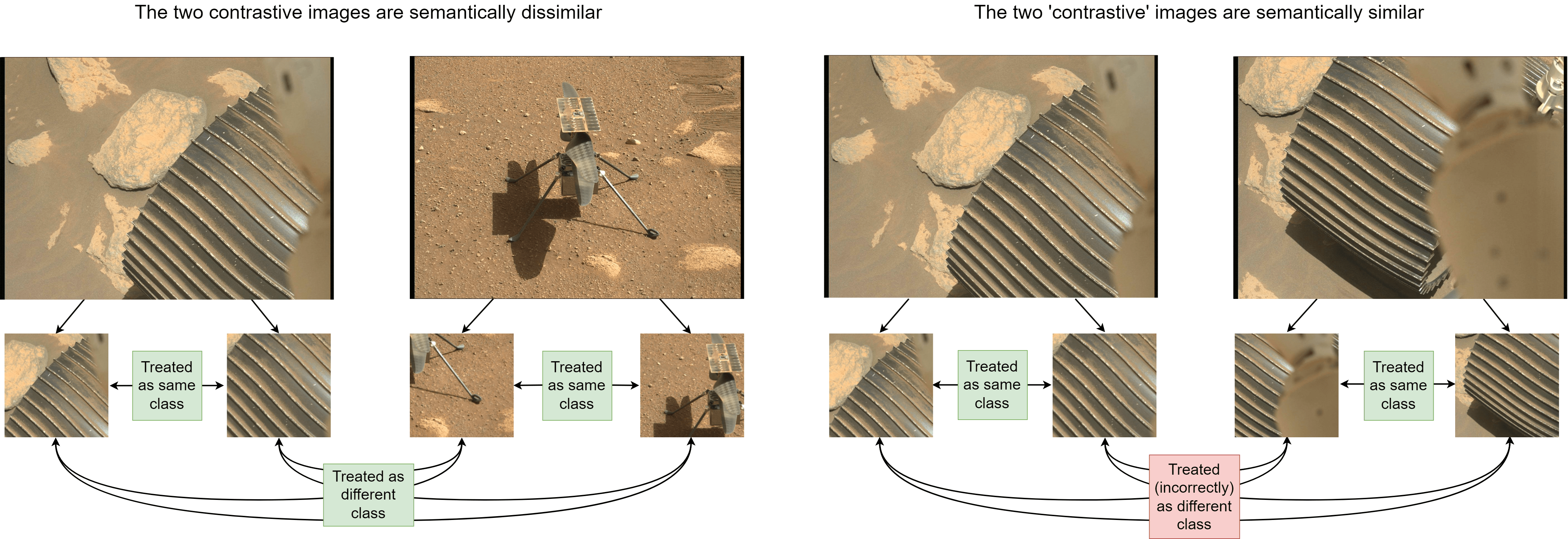}
    \caption{The contrastive assumption is being held (left), as the two sampled images \textit{are} in visual contrast. The contrastive assumption is being violated (right) as the two sampled images are not in contrast, \textit{but they are assumed to be by the baseline contrastive learning framework}. The violating pairs are thus given an incorrect pseudo-label during training, causing reduced performance. This issue is exacerbated as more semantically similar images are sampled into a training batch. }
    \label{fig:contrastive-2-cases}
\end{figure}

\subsection{Contrastive learning}

\strawman{
\begin{itemize}
    \item We are seeking to address violations of the contrastive assumption by
    \item 1) identifying them prior to training using unsupervised clustering
    \item 2) avoiding them in the first place (combining datasets)
    \item 1) entails the use of a clustering algorithm (tejas) and a modification to the loss function
    \item 2) entails the concatenation of different datasets before training
\end{itemize}
}

Our baseline method in this work is based on SimCLR(v2) \cite{chen2020simclr,chen2020simclrv2}, which is a contrastive learning framework that enables deep learning models to learn efficient representations without the need for labeled images. We introduce two approaches to address violations of the contrastive assumption (illustrated in Figure~\ref{fig:contrastive-2-cases}). The first approach modifies the SimCLR framework to use the results from an unsupervised clustering process to identify violations, and the second modification addresses the same issue by altering the distribution of the data that the model ingests.


Whereas traditional supervised learning techniques attach a manually generated annotation or label to an input image, SimCLR operates by automatically annotating pairs of images. Two views from each source image in a batch are taken and augmented. All pairs of views are analysed and labeled as a \textit{positive} pair if they came from the same source image, otherwise they are labeled as a \textit{negative} pair. This work follows SimCLR in generating augmented views of each training image by taking a random crop of the source image and applying the following transformations at random: horizontal flipping, color jittering, and grayscaling \cite{chen2020simclrv2} \footnote{We specify all hyperparameters, optimiser settings, and model configurations for each experiment in full in Appendix \ref{sec:hyperparams}.}.

A neural network then completes a forward pass over each of the views in the batch, creating a batch of embedding vectors. For any given positive pair of views, we know that their embeddings should be similar, and for any given negative pair of views, their embeddings should be dissimilar. In practice, cosine similarity is used as the similarity measure for the embedding vectors, and this defines an objective which can be optimised using gradient descent (Equation \ref{eq:contloss}), thus allowing the network to be trained.






\begin{equation}
\label{eq:contloss}
\mathcal{L}_{\textnormal{contrastive}}=-\log \frac{\exp \left(\operatorname{sim}\left(z_{i}, z_{j}\right) / \tau\right)}{\sum_{k=1}^{M} \mathbbm{1}_{[k \neq i]} \exp \left(\operatorname{sim}\left(z_{i}, z_{k}\right) / \tau\right)}
\end{equation}

\noindent Where $z_x$ refers to the $x^{th}$ view's representation, sim($u,v$) is the cosine similarity of some embedding vectors $u$ and $v$, $B$ is the batch size, $M$ is the number of views in the batch ($M = 2 \cdot B$), $\tau$ is a temperature parameter, and $i$, $j$, and $k$ are view indices. Note how identical views ($k=i$) are ignored when calculating the loss.




\subsection{Cluster-aware contrastive learning}

The assumptions of standard SimCLR fail when two views from different images actually contain the same or highly similar visual content. In this case, they will be labeled incorrectly as a negative pair; they are a \textit{false negative} (see Figure~\ref{fig:contrastive-2-cases}, right). Our first modification to SimCLR addresses this using prior information relating to if two images actually contain similar visual content.


We gather this information using the unsupervised clustering technique outlined in \cite{panambur2022self}, based on a ResNet-18 backbone. We input the unlabeled training dataset, and the output is a partition of this dataset into some manually defined number of clusters $K$. Clusters are defined by the similarity of their semantic content, with a specific focus on geological texture as a result of the deep texture encoding module that is incorporated into the architecture \cite{xue2018deep}. Since this clustering approach is unsupervised, the end-to-end contrastive training process still requires no supervision.

During training, we define positive view pairs as views which came from the same source image \textit{or cluster}. This has the effect of turning false negatives into true positives; taking view pairs that would have otherwise been incorrectly labeled during training, and converting them into correctly labeled instances. 

\begin{equation}
\label{eq:contloss-cluster}
\mathcal{L}_{\textnormal{cluster-aware}}=-\log \frac{\exp \left(\operatorname{sim}\left(z_{i}, z_{j}\right) / \tau\right)}{\sum_{k=1}^{M} \mathbbm{1}_{[c_k \neq c_i]} \exp \left(\operatorname{sim}\left(z_{i}, z_{k}\right) / \tau\right)}
\end{equation}

The modified loss function for the cluster-aware contrasive learning method is outlined in Equation \ref{eq:contloss-cluster}, where $c_x$ refers to the cluster index of the $x^{th}$ view in the batch. The key difference being that view pairs are considered negative based on their cluster index, rather than their view index.

\subsection{Mixed-domain contrastive learning}

We observe that the rate of false negative view pairs is related to the semantic homogeneity of a training dataset, and as such, we hypothesize that increasing the visual variance of the dataset should decrease the rate of false negatives encountered during contrastive learning. We achieve this by simply concatenating and shuffling two visually different datasets and then performing baseline SimCLR contrastive training on the resultant dataset. 

However, mixed-domain training may also have negative side-effects in that injecting too many out-of-domain images may ultimately reduce performance on Mars-related vision tasks. In this study, to achieve a reduced rate of false negative view pairs while preserving a sufficient amount of in-domain images, we perform mixed-domain contrastive learning by combining unlabeled Curiosity images and ImageNet images during pretraining.


\subsection{Benchmarking learned feature extraction with linear evaluation}

Both SimCLR and our modified variants take a set of unlabeled training images and train a neural network to extract discriminant features from such images. Naturally, it is our desire to quantify the success of this training paradigm, and the resultant feature extractor. To do so, we apply the contrastively-trained backbone to supervised Mars-related vision tasks, namely MSL v2.1 and HiRISE v3.2, which represent in-domain and out-of-domain challenges respectively (relative to the unlabeled training dataset which is comprised of Mars rover images).

When benchmarking, we precompute the set of features for the benchmark images using the contrastively-trained model, and then use a single dense / linear layer of $128$ neurons to learn a mapping from the features to the images' class. We train this evaluation layer using the features we extracted from the MSL v2.1 and HiRISE v3.2 training subsets and their corresponding labels, before benchmarking the performance on the datasets' corresponding testing subsets. Importantly, the parameters of the contrastively-trained model remain untouched in this process; we are only testing how useful the extracted features are when completing downstream Mars-related vision tasks. This form of linear evaluation protocol is consistent with other contrastive-learning works \cite{chen2020simclr,chen2020simclrv2,wang2022ssl4eo,he2019moco,chen2020mocov2}.

\section{Results and discussion}
\label{sec:results}

\subsection{Adhering to the contrastive assumption strengthens performance} 

\label{sec:addressing-violations}

\begin{figure}
     \centering
     \begin{subfigure}[b]{0.47\textwidth}
         \centering
         \includegraphics[width=\textwidth]{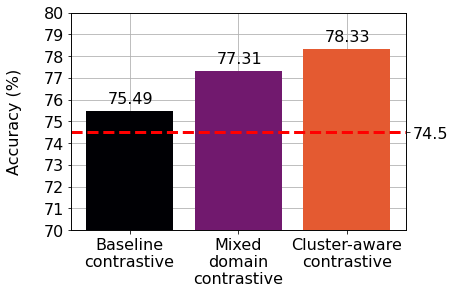}
         \caption{MSL v2.1}
         \label{fig:main-result-msl}
     \end{subfigure}
     \begin{subfigure}[b]{0.47\textwidth}
         \centering
         \includegraphics[width=\textwidth]{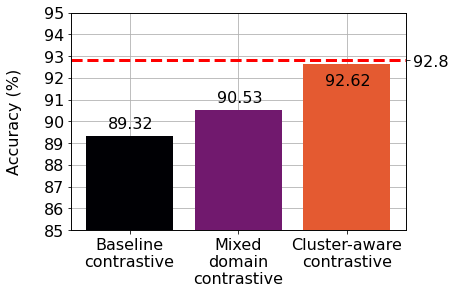}
         \caption{HiRISE v3.2}
         \label{fig:main-result-hirise}
     \end{subfigure}
        \caption{ Performance improvements with mixed-domain contrastive learning and cluster-aware contrastive learning. With a pretraining process that requires no labels, contrastive pretraining generally outperforms the supervised transfer learning baseline  (marked with a dotted line) with a model one third the size.}
        \label{fig:main-result}
\end{figure}



We began by investigating how addressing violations of the contrastive assumption improve the quality of learned features. We trained a feature extractor using the baseline SimCLR and the two modified SimCLR approaches on the unlabeled Perseverance and Curiosity datasets outlined in Section \ref{sec:datasets}, and then compared their performance on the MSL v2.1 classification benchmark, with the results outlined in Figure \ref{fig:main-result}.

Figure \ref{fig:main-result} shows that each of the contrastive learning approaches (baseline, mixed-domain, cluster-aware) exceeds Wagstaff et al.'s published baseline results (obtained using supervised transfer learning) on the MSL v2.1 benchmark \cite{wagstaff2021mars}. 

We note that other published techniques do outperform our model on the MSL v2.1 classification benchmark, but these approaches incorporate labels into a semi-supervised learning approach ($95.86\%$ test accuracy on MSL v2.1) \cite{wang2021semi} , or take a fully supervised approach with attention-based models ($81.53\%$ test accuracy on MSL v2.1) \cite{chakravarthy2021mrscatt}.


Importantly, no labels were required to train our contrastive feature extractors, whereas Wagstaff et al.'s model required a supervised dataset in addition to 1M+ labeled ImageNet images (for pretraining).
Moreover, we note that Wagstaff et al.'s fully supervised approach used a neural network backbone with approximately $60$M parameters (AlexNet \cite{krizhevsky2012alexnet}), whereas our approaches use a backbone with approximately $21$M parameters (ResNet-50 \cite{he2016deep,wightman2019timm}).


We see that the baseline contrastive approach is outperformed by the mixed-domain approach, which is then outperformed by the cluster-aware approach. A potential explanation for this is that the mixed-domain approach \textit{minimises} contrastive assumption violations, and the cluster-aware approach \textit{corrects} said violations, hence the performance difference.



We note that the performance gains associated with our proposed modified contrastive learning approaches are comparable to those reported in literature using similar approaches designed to mitigate incorrectly labeled view pairs during contrastive training. These works report accuracy increases ranging from approximately $3$--$4\%$ \cite{huynh2022boosting,chen2021incremental}. 

\subsection{Learned features generalise to out-of-domain tasks}


We extended our testing to the out-of-domain HiRISE v3.2 benchmark --- a classification task comprised entirely of orbital images of Mars, reporting the results in Figure \ref{fig:main-result-hirise}. Wagstaff et al.'s fully supervised model was trained on HiRISE data (orbital images), whereas our contrastive models were trained on Perseverance data (rover images) --- the HiRISE v3.2 task is thus out-of-domain for our contrastive approaches, but in-domain for the fully supervised technique.


We observe that the fully supervised technique outperforms our contrastive approaches, likely due to the in-domain / out-of-domain training difference. Regardless, the contrastive methods still demonstrate comparable performance to a fully supervised approach with only a linear classification head, thus illustrating the general nature of the learned feature extraction; performance is maintained even under a domain shift and without any fine tuning of the feature extractor's parameters. Put simply: our model has never encountered an orbital image of Mars during training, yet still performs competitively with a supervised model that was trained specifically on Martian orbital images.

As in Section \ref{sec:addressing-violations}, we observe a performance increase from the baseline contrastive method, to the mixed-domain method, to the cluster-aware method.

\subsection{Mixed-domain approaches increase dataset variability and performance}
\label{sec:combining}


\strawman{
\begin{itemize}
    \item Since the underlying problem is contrastive assumption violation, combining datasets solves this too!
    \item Note the proportion of false negatives per batch on average for MSL --- the problem is greatly exaggerated on our specific niche dataset when compared to image net. (Invite the reader to consider if this issue might be in their niche domain of research)
\end{itemize}
}

The `mixed domain contrastive' results in Figure \ref{fig:main-result} indicate that increasing the semantic heterogeneity of the training data by mixing datasets from different visual domains increases the performance of the trained feature extractors with respect to downstream classification tasks.

\begin{table}[h]
  \caption{Analysis of mixed-domain contrastive learning}
  \label{table:combined-datasets}
  \centering
  \begin{tabular}{ccccccc}
    \toprule
    Domain(s) / training datasets & FN per batch (\%) & MSL Acc. (\%) & HiRISE Acc. (\%)\\
    \midrule
100K Curiosity & 15.0 & 75.60 & 89.70 \\
100K ImageNet & 0.08 & 76.54 & 91.89 \\
50K Curiosity + 50K ImageNet & 0.30 & 77.31 & 90.53 \\
    \bottomrule
  \end{tabular}
\end{table} 


Table \ref{table:combined-datasets} quantifies this effect; each row represents a different training set (or combination of training sets) that were used during training. The first two rows represent baseline contrastive learning, and the last row represents \textit{mixed-domain} contrastive learning --- as the training dataset is a mix of two domains. The \textit{number} of training images remains constant in all cases. More details on this simulation are provided in Appendix~\ref{sec:simresults}. We expect that performance will be worst when the model is trained on pure datasets that are highly semantically homogeneous (e.g. 100,000 Curiosity images). We also expect the inverse to be true --- diverse datasets (e.g. ImageNet) will result in a more performant model, as the contrastive assumption will be violated less often during training.

Moreover, we expect to find a trade off in performance; training on data that is similar to the benchmark data will result in increased performance with respect to that benchmark, but performance will still be hampered due to semantic homogeneity. A mix of two domains, one which shares the domain of the benchmark dataset, and one which provides semantic heterogeneity, should then result in increased performance. 

Column two of Table \ref{table:combined-datasets} shows the average percentage of false negatives view pairs out of all the view pairs encountered in a given training batch. These results are derived from a simulation which samples multiple batches and checks for FNs using labels. Note that the more diverse dataset (ImageNet) has a far lower false negative view pairs rate than the less diverse Curiosity dataset, and mixing these datasets 50-50 results in a FN rate equal to a `diversity-weighted' average of the two constituent datasets (ImageNet is far more diverse than the Curiosity images, with 1000 classes and 19 classes respectively).

Column three shows our results on the MSL v2.1 benchmark. We note that the model trained on pure Curiosity data (i.e., the same domain as the benchmark dataset) is the \textit{least} performant model. Training on pure ImageNet increases the performance further, and training on an equal mix of both domains (i.e., mixed-domain contrastive learning) provides the best performance.

In column four, we see equivalent results, but on a benchmark domain which is \textit{not} represented in any training dataset (HiRISE v3.2's orbital images). In this case, the benefit of having in-domain training images (i.e. Curiosity images) is removed; the most semantically heterogenous training dataset (ImageNet) provides the best performance and out-of-domain generalisation. In fact, using any Curiosity images in this case \textit{reduces performance}. This raises a useful conclusion: if no data from the target testing domain is available, use the most semantically heterogeneous dataset available (when using contrastive learning). If data from the target testing domain is available, then using a mix of training data may provide increased performance.

\section{Ablation studies}

\subsection{$10\%$ of the labels is sufficient to transfer to other tasks}

\begin{figure}[h]
    \centering
    \begin{subfigure}[b]{0.45\textwidth}
        \centering
        \includegraphics[width=\textwidth]{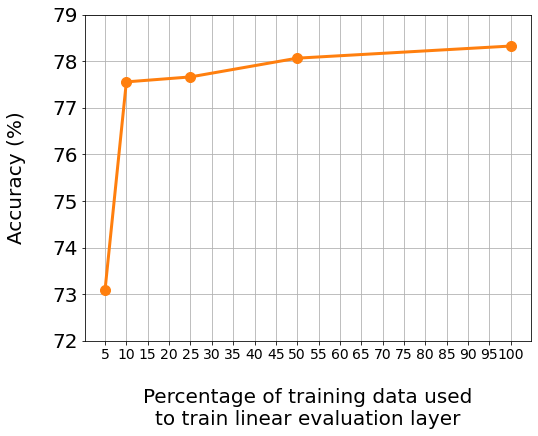}
        \caption{MSL v2.1}
    \end{subfigure}
    \begin{subfigure}[b]{0.48\textwidth}
        \centering
        \includegraphics[width=\textwidth]{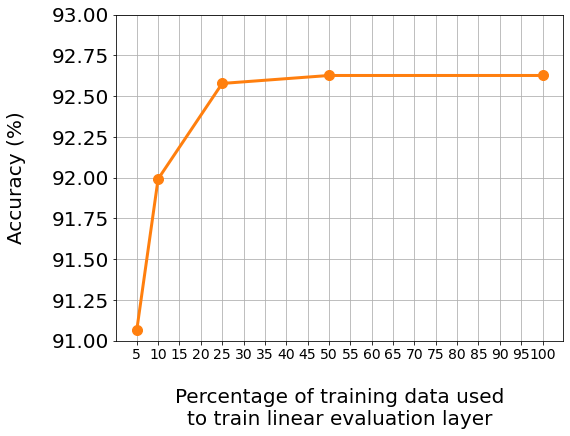}
        \caption{HiRISE v3.2}
    \end{subfigure}
    \caption{The task performance as a function of different percentages of training data used during linear evaluation. In all cases, we pretrain on the full unlabeled Perseverance dataset using our proposed cluster-aware contrastive learning, then perform linear evaluation with limited task-specific training data, then benchmark and report the results on the task's testing set. We note that the accuracy return for increasing amounts of training data quickly diminishes.}
    \label{fig:percentage-of-data}
\end{figure}

\strawman{
\begin{itemize}
    \item Using only 10\% of the labels preserves performance and shows that it doesn't take much to learn how to discriminate the features from the contrastively trained feature extractors 
    \item Shows that we don't need to manually label a heap of data to get comparable performance, even in a niche dataset like ours
    \item Discussion: the feature extractors have done the bulk of the learning on pure image data. We can tune them for our specific downstream tasks with only a little data (kinda transfer learning?)
\end{itemize}
}


Expanding on the results outlined in Section \ref{sec:number-of-clusters}, we observe in Figure \ref{fig:percentage-of-data} that performance remains relatively unchanged even when a fractional amount of the benchmark training data is used. On the MSL v2.1 benchmark, accuracy is within $1\%$ of the best score even when only $10\%$ of the task-specific training data is being used, though a distinct performance drop is noted when this proportion is reduced to $5\%$. On the HiRISE v3.2 benchmark, accuracy is within $0.5\%$ of the best score even at $25\%$ of the task-specific training data. The relatively higher proportion of training data required on the out-of-domain HiRISE v3.2 classification task may suggest that more training data is required to specialise to an out-of-domain task.

Overall, these results suggest that the majority of the learning has occurred during contrastive training and without labels, necessitating only a small number of task-specific labels to leverage the features extracted by the contrastively-trained neural network. This demonstrates key benefits of contrastive learning; the resulting models are relatively general and can be optimised for specific tasks with only a limited number of training samples (especially when compared to a fully supervised training pipeline).



\subsection{Cluster-aware contrastive learning is sensitive to the number of clusters chosen}

\label{sec:number-of-clusters}

\strawman{
\begin{itemize}
    \item The clustering approach is not always more performant --- the correct number of clusters needs to be chosen
    \item Failing to do this can decrease performance when compared to baseline contrastive learning
    \item Discussion: this may occur as it may exaggerate or not deal with the violations of the contrastive assumption appropriately
    \item This matters more when benchmarking in-domain testing datasets (in-domain when compared to the training set). The out-of-domain orbital image data shows similar performance gains across the board
    \item Discussion: this is perhaps because the semantic clusters/classes encountered during training are never relevant to the benchmark dataset, and are thus irrelevant to downstream task performance
\end{itemize}
}

We altered the number of clusters (K) during the unsupervised clustering stage of our cluster-aware contrastive approach and benchmarked the results. 


\begin{figure}[H]
    \centering
    \begin{subfigure}[b]{0.45\textwidth}
        \centering
        \includegraphics[width=\textwidth]{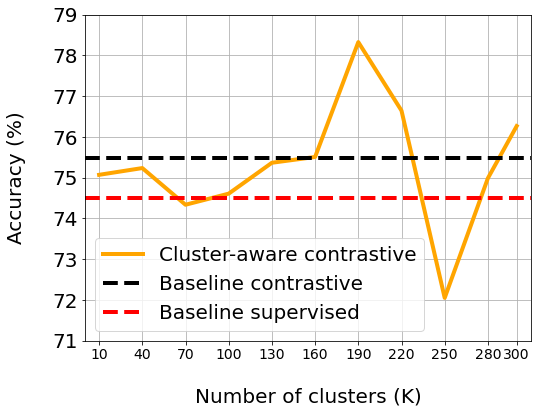}
        \caption{MSL v2.1}
    \end{subfigure}
    \begin{subfigure}[b]{0.45\textwidth}
        \centering
        \includegraphics[width=\textwidth]{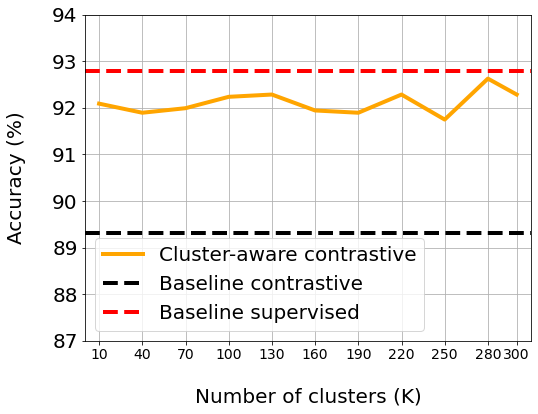}
        \caption{HiRISE v3.2}
    \end{subfigure}
    \caption{The performance of the contrastively-trained feature extractors, as measured with respect to an in-domain (left) and out-of-domain (right) downstream task, is dependent on the number of clusters computed prior to training. Too few clusters and images that are dissimilar in content may be assigned to the same cluster, too many clusters and images that are similar in content may be assigned to different clusters --- both will cause incorrect pseudo-labels during contrastive training.}
    \label{fig:k-sweep}
\end{figure}

We note that the hyperparameter K requires careful optimisation, otherwise the cluster-aware method may not improve (and may decrease) performance. We posit that too few clusters may fail to adequately address violations of the contrastive assumption; images with different semantic content may be placed into the same clusters due to the coarseness of the clustering, thus failing to solve the issue at hand. With too many clusters, the effectiveness of the method may diminish, with only weakly contrasting images being labeled as negatives.

We observe that for the out-of-domain task HiRISE v3.2, the tuning of the hyperparameter K is less important, with any choice of K producing meaningful performance gains when compared to the baseline contrastive case. We hypothesize that our modified contrastive methods only attend to features inherent to the types of images that were encountered during contrastive training, thus resulting in nuances of orbital images being ignored. 

\section{Conclusion}

\strawman{
\begin{itemize}
    \item Contrastive learning can be applied even when images are not contrasting
    \item There are ways to do this without supervision --- the end to end process remains self-supervised / unsupervised
    \item Doing this thus requires no extra labeling effort and improves performance when compared to baseline contrastive learning
    \item It also improves performance when compared to baseline supervised learning
    \item The approaches outlined in this paper are promising for the domain of space-related datasets, which are often composed of many similar images, and thus violate the central assumption of contrastive learning --- that any two given images from a dataset will generally contain contrasting semantic content
    \item In general these techniques can be used to improve performance and train generalised feature extractors in datasets with low semantic variance among images
\end{itemize}
}

In this work, we have explored and quantified a common issue native to contrastive learning on semantically homogeneous datasets; violations of the contrastive assumption leading to false negative view pairs. We analysed this issue within the purview of Mars images taken from the rovers Curiosity and Perseverance, and from the Mars Reconnaissance Orbiter, and we proposed two modifications to contrastive learning tailored for this domain. 

Our experiments demonstrate how these modifications --- cluster-aware and mixed-domain contrastive learning --- improve upon baseline contrastive learning by training more discriminant and powerful feature extractors. In each case, we compare our results to a fully supervised baseline, and note that for in-domain classification tasks our methods result in feature extractors that exceed the performance of their fully supervised counterparts, even when using only $10\%$ of the available labels. It is our hope that this work illustrates how the benefits of contrastive learning can be applied to the large scale space datasets --- even when the underlying visual data is semantically homogeneous or not sufficiently contrastive --- thus increasing the scope for contrastive learning in the domain of space images.





\section{Acknowledgments}

This research was carried out at the Jet Propulsion Laboratory, California Institute of Technology, under a contract with the National Aeronautics and Space Administration (80NM0018D0004), and was funded by the Data Science Working Group (DSWG). The authors also acknowledge the Extreme Science and Engineering Discovery Environment (XSEDE) Bridges at Pittsburgh Supercomputing Center for providing GPU resources through allocation TG-CIS220027. U.S. Government sponsorship acknowledged.

%
%
\bibliographystyle{splncs04}
\bibliography{egbib}

\begin{thebibliography}{10}
\providecommand{\url}[1]{\texttt{#1}}
\providecommand{\urlprefix}{URL }
\providecommand{\doi}[1]{https://doi.org/#1}

\bibitem{sklearn-api}
Buitinck, L., Louppe, G., Blondel, M., Pedregosa, F., Mueller, A., Grisel, O.,
  Niculae, V., Prettenhofer, P., Gramfort, A., Grobler, J., Layton, R.,
  VanderPlas, J., Joly, A., Holt, B., Varoquaux, G.: {API} design for machine
  learning software: experiences from the scikit-learn project. In: ECML PKDD
  Workshop: Languages for Data Mining and Machine Learning. pp. 108--122 (2013)

\bibitem{caron2021emerging}
Caron, M., Touvron, H., Misra, I., J\'egou, H., Mairal, J., Bojanowski, P.,
  Joulin, A.: Emerging properties in self-supervised vision transformers. In:
  Proceedings of the International Conference on Computer Vision (ICCV) (2021)

\bibitem{chakravarthy2021mrscatt}
Chakravarthy, A.S., Roy, R., Ravirathinam, P.: Mrscatt: A spatio-channel
  attention-guided network for mars rover image classification. In: Proceedings
  of the IEEE/CVF Conference on Computer Vision and Pattern Recognition. pp.
  1961--1970 (2021)

\bibitem{chen2020simclr}
Chen, T., Kornblith, S., Norouzi, M., Hinton, G.: A simple framework for
  contrastive learning of visual representations. In: International conference
  on machine learning. pp. 1597--1607. PMLR (2020)

\bibitem{chen2020simclrv2}
Chen, T., Kornblith, S., Swersky, K., Norouzi, M., Hinton, G.: Big
  self-supervised models are strong semi-supervised learners. arXiv preprint
  arXiv:2006.10029  (2020)

\bibitem{chen2021incremental}
Chen, T.S., Hung, W.C., Tseng, H.Y., Chien, S.Y., Yang, M.H.: Incremental false
  negative detection for contrastive learning. arXiv preprint arXiv:2106.03719
  (2021)

\bibitem{chen2020mocov2}
Chen, X., Fan, H., Girshick, R., He, K.: Improved baselines with momentum
  contrastive learning. arXiv preprint arXiv:2003.04297  (2020)

\bibitem{dosovitskiy2020image}
Dosovitskiy, A., Beyer, L., Kolesnikov, A., Weissenborn, D., Zhai, X.,
  Unterthiner, T., Dehghani, M., Minderer, M., Heigold, G., Gelly, S., et~al.:
  An image is worth 16x16 words: Transformers for image recognition at scale.
  arXiv preprint arXiv:2010.11929  (2020)

\bibitem{goh2022mars}
Goh, E., Chen, J., Wilson, B.: Mars terrain segmentation with less labels.
  arXiv preprint arXiv:2202.00791  (2022)

\bibitem{he2019moco}
He, K., Fan, H., Wu, Y., Xie, S., Girshick, R.: Momentum contrast for
  unsupervised visual representation learning. arXiv preprint arXiv:1911.05722
  (2019)

\bibitem{he2016deep}
He, K., Zhang, X., Ren, S., Sun, J.: Deep residual learning for image
  recognition. In: Proceedings of the IEEE conference on computer vision and
  pattern recognition. pp. 770--778 (2016)

\bibitem{helber2019eurosat}
Helber, P., Bischke, B., Dengel, A., Borth, D.: Eurosat: A novel dataset and
  deep learning benchmark for land use and land cover classification. IEEE
  Journal of Selected Topics in Applied Earth Observations and Remote Sensing
  \textbf{12}(7),  2217--2226 (2019)

\bibitem{huynh2022boosting}
Huynh, T., Kornblith, S., Walter, M.R., Maire, M., Khademi, M.: Boosting
  contrastive self-supervised learning with false negative cancellation. In:
  Proceedings of the IEEE/CVF Winter Conference on Applications of Computer
  Vision. pp. 2785--2795 (2022)

\bibitem{krizhevsky2012alexnet}
Krizhevsky, A., Sutskever, I., Hinton, G.E.: Imagenet classification with deep
  convolutional neural networks. Advances in neural information processing
  systems  \textbf{25} (2012)

\bibitem{kucik2021snn}
Kucik, A.S., Meoni, G.: Investigating spiking neural networks for
  energy-efficient on-board ai applications. a case study in land cover and
  land use classification. In: Proceedings of the IEEE/CVF Conference on
  Computer Vision and Pattern Recognition. pp. 2020--2030 (2021)

\bibitem{van2008tsne}
Van~der Maaten, L., Hinton, G.: Visualizing data using t-sne. Journal of
  machine learning research  \textbf{9}(11) (2008)

\bibitem{nasa2022marsgoals}
NASA: Mars exploration rover mission goals,
  \url{https://mars.nasa.gov/mer/mission/science/goals/}

\bibitem{panambur2022self}
Panambur, T., Chakraborty, D., Meyer, M., Milliken, R., Learned-Miller, E.,
  Parente, M.: Self-supervised learning to guide scientifically relevant
  categorization of martian terrain images. arXiv preprint arXiv:2204.09854
  (2022)

\bibitem{paszke2019pytorch}
Paszke, A., Gross, S., Massa, F., Lerer, A., Bradbury, J., Chanan, G., Killeen,
  T., Lin, Z., Gimelshein, N., Antiga, L., Desmaison, A., Kopf, A., Yang, E.,
  DeVito, Z., Raison, M., Tejani, A., Chilamkurthy, S., Steiner, B., Fang, L.,
  Bai, J., Chintala, S.: Pytorch: An imperative style, high-performance deep
  learning library. In: Advances in Neural Information Processing Systems 32,
  pp. 8024--8035. Curran Associates, Inc. (2019),
  \url{http://papers.neurips.cc/paper/9015-pytorch-an-imperative-style-high-performance-deep-learning-library.pdf}

\bibitem{scikit-learn}
Pedregosa, F., Varoquaux, G., Gramfort, A., Michel, V., Thirion, B., Grisel,
  O., Blondel, M., Prettenhofer, P., Weiss, R., Dubourg, V., Vanderplas, J.,
  Passos, A., Cournapeau, D., Brucher, M., Perrot, M., Duchesnay, E.:
  Scikit-learn: Machine learning in {P}ython. Journal of Machine Learning
  Research  \textbf{12},  2825--2830 (2011)

\bibitem{russakovsky2015imagenet}
Russakovsky, O., Deng, J., Su, H., Krause, J., Satheesh, S., Ma, S., Huang, Z.,
  Karpathy, A., Khosla, A., Bernstein, M., et~al.: Imagenet large scale visual
  recognition challenge. International journal of computer vision
  \textbf{115}(3),  211--252 (2015)

\bibitem{sprinks2016mars}
Sprinks, J.C., Wardlaw, J., Houghton, R., Bamford, S., Marsh, S.: Mars in
  motion: An online citizen science platform looking for changes on the surface
  of mars. In: AAS/Division for Planetary Sciences Meeting Abstracts\# 48.
  vol.~48, pp. 426--01 (2016)

\bibitem{swan2021ai4mars}
Swan, R.M., Atha, D., Leopold, H.A., Gildner, M., Oij, S., Chiu, C., Ono, M.:
  Ai4mars: A dataset for terrain-aware autonomous driving on mars. In:
  Proceedings of the IEEE/CVF Conference on Computer Vision and Pattern
  Recognition. pp. 1982--1991 (2021)

\bibitem{wagstaff2021mars}
Wagstaff, K., Lu, S., Dunkel, E., Grimes, K., Zhao, B., Cai, J., Cole, S.B.,
  Doran, G., Francis, R., Lee, J., et~al.: Mars image content classification:
  Three years of nasa deployment and recent advances. arXiv preprint
  arXiv:2102.05011  (2021)

\bibitem{wagstaff2018deep}
Wagstaff, K.L., Lu, Y., Stanboli, A., Grimes, K., Gowda, T., Padams, J.: Deep
  mars: Cnn classification of mars imagery for the pds imaging atlas. In:
  Thirty-Second AAAI Conference on Artificial Intelligence (2018)

\bibitem{wang2021semi}
Wang, W., Lin, L., Fan, Z., Liu, J.: Semi-supervised learning for mars imagery
  classification. In: 2021 IEEE International Conference on Image Processing
  (ICIP). pp. 499--503. IEEE (2021)

\bibitem{wang2022ssl4eo}
Wang, Y., Braham, N.A.A.A., Albrecht, C.M., Xiong, Z., Liu, C., Zhu, X.X.:
  Ssl4eo-s12: A large-scale multimodal multitemporal dataset for
  self-supervised learning in earth observation  (2022)

\bibitem{wightman2019timm}
Wightman, R.: Pytorch image models.
  \url{https://github.com/rwightman/pytorch-image-models} (2019).
  \doi{10.5281/zenodo.4414861}

\bibitem{xue2018deep}
Xue, J., Zhang, H., Dana, K.: Deep texture manifold for ground terrain
  recognition. In: Proceedings of the IEEE Conference on Computer Vision and
  Pattern Recognition. pp. 558--567 (2018)

\bibitem{zagoruyko2016wide}
Zagoruyko, S., Komodakis, N.: Wide residual networks. arXiv preprint
  arXiv:1605.07146  (2016)

\end{thebibliography}

\appendix

\section{Hyperparameter configurations}

\label{sec:hyperparams}

\begin{table}[H]
    \centering
    \setlength{\tabcolsep}{8pt}
    \caption{ The default hyperparameters used in all contrastive training / linear evaluation experiments. All experiments are executed in PyTorch \cite{paszke2019pytorch}.}
    
    
    \begin{tabular}{l l}
        \hline
        Hyperparameter  & Value(s) \\
        \hline
        
        GPUs & 4x NVIDIA Tesla V100 \\
        Training type & Distributed data parallel \\
        CPU workers & 16 \\
        model backbone & ResNet-50 (vanilla) \\
        model width \cite{zagoruyko2016wide} & 1x \\
        image size & $224$ \\
        num. epochs & $400$ \\
        num. projection layers & $3$ \\
        projection layer dimension & $128$ \\
        linear eval. dimension & $128$ \\
        optimiser & Adam \\
        learning rate & $3e$-$4$\\
        weight decay & $1e$-$6$ \\
        temperature & $0.5$ \\
        batch size & $128$ \\ 
        
        \hline
    \end{tabular}
    \label{tab:contrastive-params}
\end{table}

\section{Investigating the relationship between semantic homogeneity and false negatives}

\label{sec:simresults}

\strawman{
\begin{itemize}
    \item 1) Changing the dataset size makes no difference to the individual batch, beyond a point of triviality
    \item 2) Increase the batch size does not greatly change the false negative rate (y axis). If the underlying distribution of classes in the dataset is unchanging, then any given batch will still sample the same rate of false negative source images
    \item 3) The semantic variation/classes in a dataset does impact the false negative proportion in a given batch. We analyse the datasets themselves in the next section \todo{(I could show the false negative proprtion as a function of number of clusters!)}. \todo{Wait no, that would just decrease to zero, it wouldn't peak. I would also need to have some measure of the maximum contrast between images, where they intersect is the ideal number of clusters}
\end{itemize}
}

We investigate how the rate of false negative view pairs in a batch is a function of dataset’s semantic homogeneity. We start by simulating the sampling of one thousand batches from datasets with uniformly distributed class frequencies for the purpose of characterising the false negative view pair problem with respect to training and dataset parameters. In each sample, we compute all view pairs and report the average proportion of false negative view pairs amongst all view pairs in that batch. The more false negative view pairs, the worse the expected performance of the final model.

We find that --- beyond an initial point --- the percentage of false negative view pairs in any given training batch is invariant to the size of the dataset and the size of the batch. Changing the dataset's size naturally makes no difference, as any given batch will still sample images belonging to any given semantic class with the same frequency. Changing the batch size simply results in proportionally more false negative view pairs being sampled, so again, this does not address the problem.

The false negative rate however does decrease from $10\%$ asymptotically to $0\%$ as the number of semantic classes (i.e., the variation in visual content) grows from $0$ to $1000$ distinct classes. We can see how violations of the contrastive assumption disproporionately affect smaller, less variable datasets (e.g., between $0$ and $50$ classes) relative to larger, more variable datasets (e.g. ImageNet). This explains why combining different datasets minimises violations of the contrastive assumption, false negative view pairs, and thus increases performance on less visually diverse datasets (e.g. Mars rover image datasets).

Importantly, the Mars image datasets are highly imbalanced (see Section \ref{sec:datasets}), thus exaggerating the issues outlined here. In Section~\ref{sec:combining}, we perform similar simulations, but tailor them for specific datasets (e.g. ImageNet and the Curiosity dataset). For ImageNet, we can determine if a false negative would appear in a batch by consulting the labels, and for the Curiosity dataset, we simulate using the MSL v2.1 class distribution, which we assume to be representative as it was derived from the Curiosity dataset.  

\begin{figure}[h]
    \centering
    \includegraphics[width=0.9\textwidth]{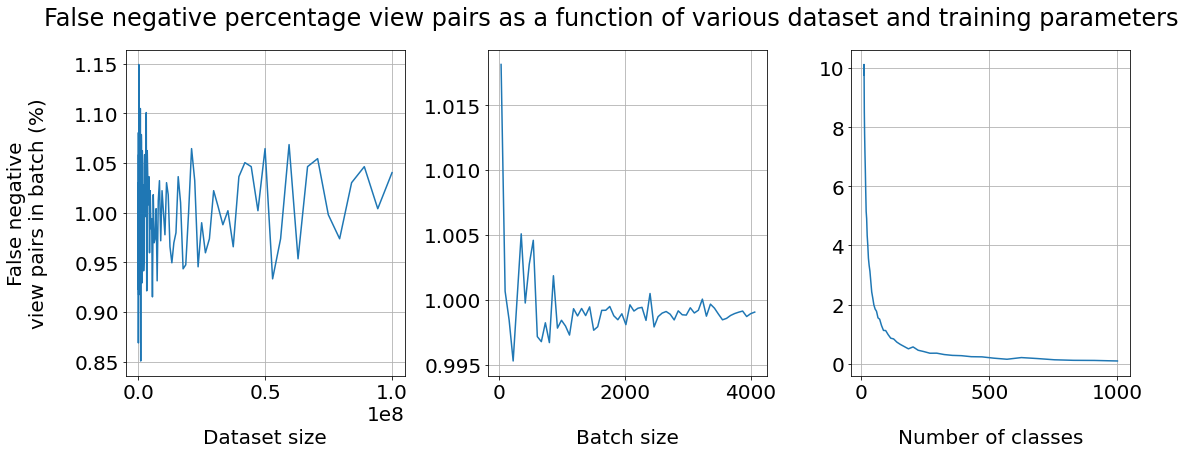}
    
    \caption{Simulation results illustrating how the proportion of false negative pairs in a batch change with changing dataset statistics and hyperparameters. These simulation results are for a theoretical dataset with a perfectly balanced class distribution.} 
    \label{fig:fn-sims}
    
\end{figure}

\section{Visualising the assignment of pseudo-labels}

\begin{figure}[H]
    \centering
    \includegraphics[width=0.9\textwidth]{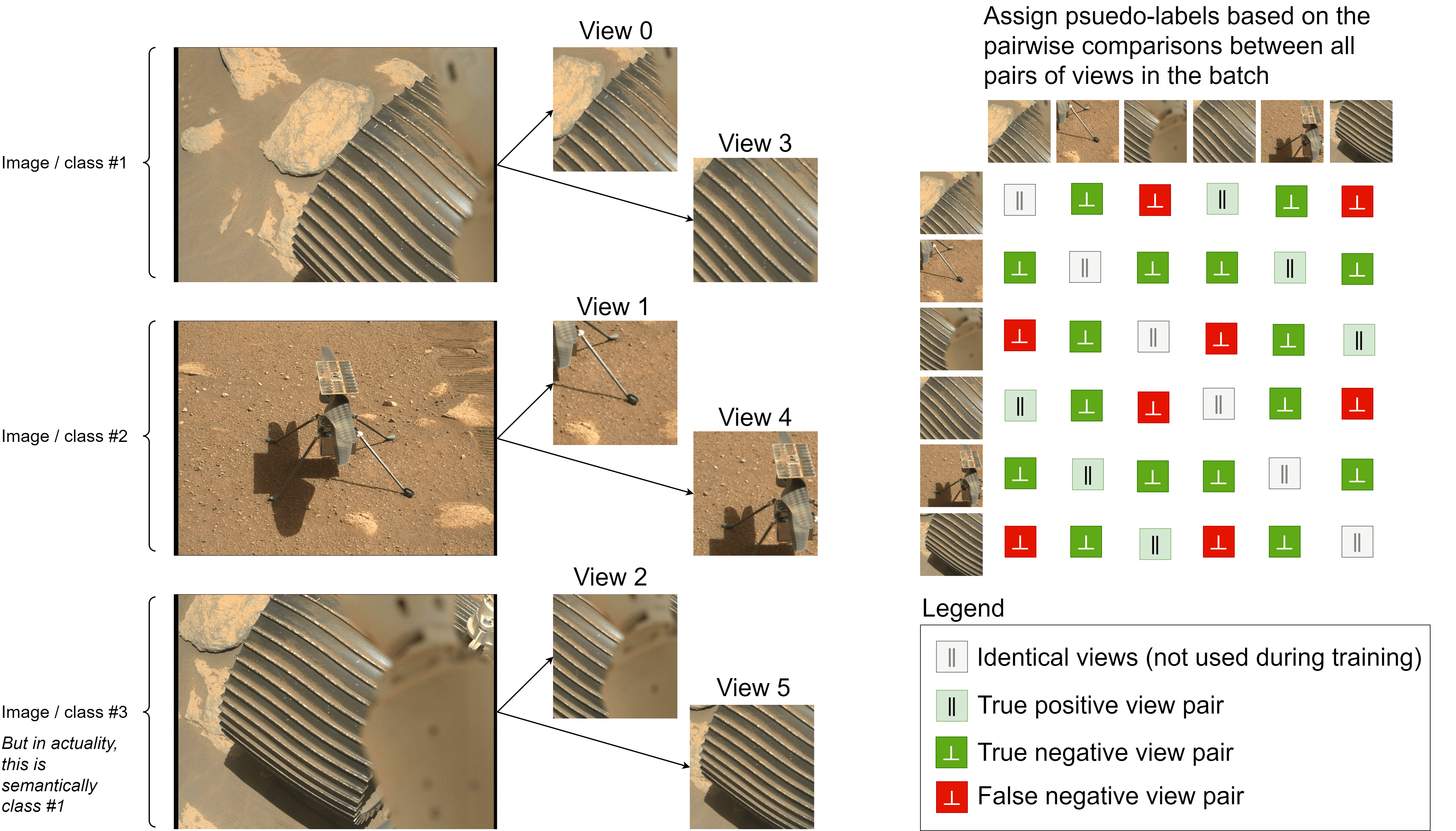}
    
    \caption{We visualise the assignment of true positive, true negative, and false negative view pairs for an example batch of $3$ source images. This batch violates the contrastive assumption, as images $1$ and $3$ feature images with the same semantic content, and will thus generate false negative view pairs. Each image has $2$ views extracted from it, and every pair of views is considered as a training instance in the batch (apart from views paired with themselves, i.e., the main diagonal of the psuedo-label matrix shown on the right). Note that when two views from image \#1 and image \#3 are paired, they will falsely be considered as negatives (e.g. view 0 and view 5, which actually feature the same semantic content). }
    \label{fig:batch-psuedolabels}
    
\end{figure}

The process by which pseudo-labels are assigned to each pair of views in baseline, unmodified contrastive learning is visualised in Figure~\ref{fig:batch-psuedolabels}.

\section{t-SNE clustering visualisations of Perseverance data}

\begin{figure}[H]
    \centering
    \includegraphics[width=0.9\textwidth]{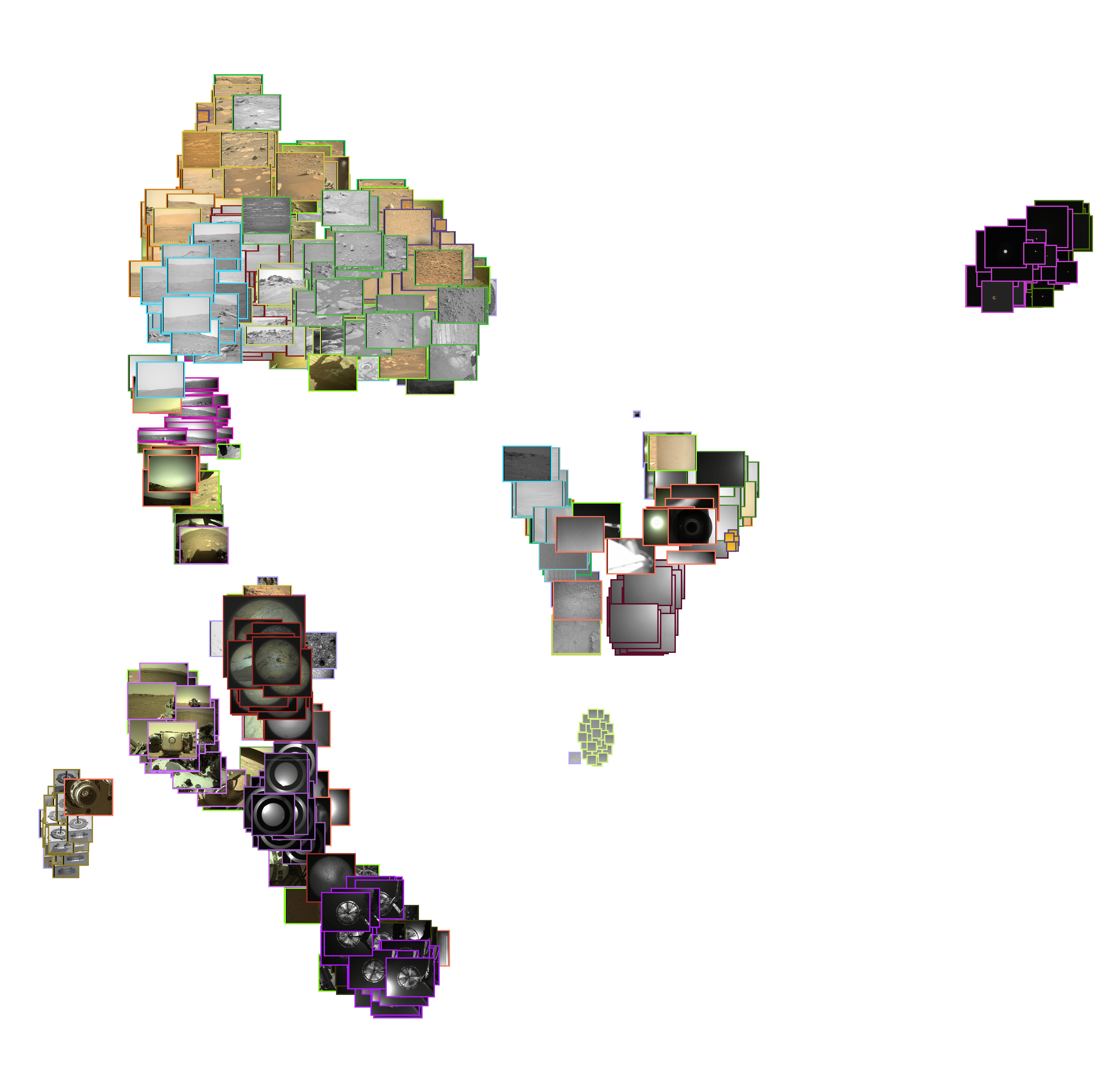}
    
    \caption{The cluster-aware contrastive model generates features which are clustered by their semantic content. Distinct clusters of images are grouped due to their feature similarity; the rocket-powered sky crane images are clustered in purple, Perseverance's sundial images are clustered in yellow, as are images with similar landscape image content, aspect ratios, colour profiles, vignetting, and so on. }
    \label{fig:tsne}
    
\end{figure}

We input the hidden features for the Perseverance dataset generated by our cluster-aware contrastive models into the t-distributed stochastic neighbor embedding (t-SNE) method \cite{van2008tsne} provided by scikit-learn \cite{scikit-learn,sklearn-api}. This statistical method reduces the embeddings to 2-dimensions whilst problematically ensuring that similar features remain close and disimilar features remain far away in the visualisation. At each point in the visualisation (Figure~\ref{fig:tsne}), we plot the hidden feature's source image, and we consider $25$ classes for the purpose of visualisation.

\end{document}